\documentclass{article}
\usepackage[utf8]{inputenc} 
\usepackage[T1]{fontenc}    
\usepackage{hyperref}       
\usepackage{url}            
\usepackage{booktabs}       
\usepackage{amsfonts}       
\usepackage{nicefrac}       
\usepackage{microtype}      
\usepackage{xcolor}         
\usepackage{graphicx}

\usepackage{algorithmic}
\usepackage{booktabs}
\usepackage{amsmath}
\usepackage{amssymb}
\usepackage{mathtools}
\usepackage{amsthm}
\usepackage{times}
\usepackage{multirow}
\usepackage{amsfonts}
\usepackage{import}
\usepackage{pifont}
\usepackage{textcomp}
\usepackage{color, colortbl}
\usepackage{xcolor}
\definecolor{greyC}{RGB}{180,180,180}
\definecolor{greyL}{RGB}{235,235,235}
\usepackage{footmisc}
\usepackage{bm}
\usepackage[flushleft]{threeparttable}
\usepackage{makecell}
\usepackage{hyperref}
\usepackage{tcolorbox}
\usepackage{array}     
\usepackage{tabularx}
\usepackage{xcolor}
\usepackage{multirow}
\usepackage{booktabs}
\usepackage{makecell}

\usepackage{hyperref}
\usepackage{url}
\usepackage{enumitem}
\usepackage{tabularx}
\usepackage{subcaption}
\usepackage{wrapfig}
\usepackage{amsmath}
\usepackage{amssymb}
\usepackage{mathtools}
\usepackage{amsthm}

\usepackage{algorithm}

\usepackage[utf8]{inputenc}
\usepackage{graphicx}
\usepackage{amsmath}
\usepackage{amsthm}
\usepackage{booktabs}
\usepackage{pifont}
\usepackage{array} 

\definecolor{deepblue}{RGB}{180,220,250}    
\definecolor{lightblue}{RGB}{220,235,250}   

\definecolor{deeppink}{RGB}{250,200,220}    
\definecolor{lightpink}{RGB}{250,220,235}   
\definecolor{mygreen}{RGB}{34,139,34} 
\definecolor{lightyellow}{RGB}{255,250,240} 
\definecolor{lightgreen}{RGB}{240,250,240}  
\definecolor{champagne}{RGB}{247, 231, 206} 
\definecolor{darksalmon}{rgb}{0.91, 0.59, 0.48}
\definecolor{lightpurple}{RGB}{235,223,249}

\usepackage{xcolor}

\definecolor{cmnt}{RGB}{0,128,128} 

\definecolor{mycomment}{RGB}{0,128,128} 

\DeclareMathAlphabet{\mathsfit}{\encodingdefault}{\sfdefault}{m}{sl}
\SetMathAlphabet{\mathsfit}{bold}{\encodingdefault}{\sfdefault}{bx}{n}

\ifx\BlackBox\undefined
\newcommand{\BlackBox}{\rule{1.5ex}{1.5ex}}  
\fi

\ifx\QED\undefined
\def\QED{~\rule[-1pt]{5pt}{5pt}\par\medskip}
\fi

\ifx\proof\undefined
\newenvironment{proof}{\par\noindent{\em Proof:\ }}{\hfill\BlackBox\\[.0mm]}
\fi

\ifx\theorem\undefined
\newtheorem{theorem}{Theorem}[section]
\fi

\ifx\example\undefined
\newtheorem{example}{Example}[section]
\fi

\ifx\property\undefined
\newtheorem{property}{Property}
\fi

\ifx\lemma\undefined
\newtheorem{lemma}{Lemma}[section]
\fi

\ifx\proposition\undefined
\newtheorem{proposition}{Proposition}[section]
\fi

\ifx\remark\undefined
\newtheorem{remark}{Remark}
\fi

\ifx\corollary\undefined
\newtheorem{corollary}{Corollary}[section]
\fi

\ifx\definition\undefined
\newtheorem{definition}{Definition}[section]
\fi

\ifx\conjecture\undefined
\newtheorem{conjecture}{Conjecture}[section]
\fi

\ifx\axiom\undefined
\newtheorem{axiom}[theorem]{Axiom}
\fi

\ifx\claim\undefined
\newtheorem{claim}[theorem]{Claim}
\fi

\ifx\assumption\undefined
\newtheorem{assumption}{Assumption}
\fi

\ifx\problem\undefined
\newtheorem{problem}{Problem}[section]
\fi

\ifx\fact\undefined
\newtheorem{fact}{Fact}
\fi

\newcommand{\benr}{\begin{eqnarray}}
\newcommand{\eenr}{\end{eqnarray}}
\newcommand{\benrr}{\begin{eqnarray*}}
\newcommand{\eenrr}{\end{eqnarray*}}
\newcommand{\ben}{\begin{equation}}
\newcommand{\een}{\end{equation}}
\newcommand{\benn}{\begin{equation*}}
\newcommand{\eenn}{\end{equation*}}

\usepackage{amsthm}
\theoremstyle{plain}

\usepackage{comment}
\usepackage[utf8]{inputenc} 
\usepackage[T1]{fontenc}    
\usepackage{hyperref}       
\usepackage{url}            
\usepackage{booktabs}       
\usepackage{amsfonts}       
\usepackage{nicefrac}       
\usepackage{microtype}      
\usepackage{xcolor}         
\usepackage{multirow} 
\usepackage{graphicx}
\usepackage{booktabs}
\usepackage{tabularx}

\usepackage{varwidth}
\usepackage{tikz}
\usetikzlibrary{calc}
\usetikzlibrary{positioning, shapes, fit, backgrounds, decorations.pathreplacing}

\usepackage{changepage}
\usepackage{sidecap}

\definecolor{mm}{HTML}{569480}

\definecolor{mblue}{RGB}{0, 61, 124}
\definecolor{myellow}{RGB}{239, 124, 0}
\definecolor{mnavy}{RGB}{0,0,128}
\definecolor{minc}{RGB}{0,128,0}
\definecolor{mdec}{RGB}{255,0,0}
\definecolor{mhold}{RGB}{128,128,128}

\newcommand{\equal}{Equal contribution}

\usepackage{pifont}

\usepackage{acl2025}

\title{LLM-based Human Simulations Have Not Yet Been Reliable}

\author{
Qian Wang\textsuperscript{1}\thanks{Equal contribution}, 
Jiaying Wu\textsuperscript{1}\footnotemark[1], 
Zichen Jiang\textsuperscript{1}, 
Zhenheng Tang\textsuperscript{2}, \\ 
\textbf{Bingqiao Luo\textsuperscript{1},}
\textbf{Nuo Chen\textsuperscript{1},}
\textbf{Wei Chen\textsuperscript{1},}
\textbf{Huacan Wang\textsuperscript{3},}
\textbf{Bingsheng He\textsuperscript{1}} \\
\textsuperscript{1}National University of Singapore \quad
\textsuperscript{2}The Hong Kong University of Science and Technology \\
\textsuperscript{3}University of Chinese Academy of Sciences
}

\begin{document}
\maketitle

\begin{abstract}
Large Language Models (LLMs) are increasingly employed for simulating human behaviors across diverse domains. However, our position is that current LLM-based human simulations remain insufficiently reliable, as evidenced by significant discrepancies between their outcomes and authentic human actions. Our investigation begins with a systematic review of LLM-based human simulations in social, economic, policy, and psychological contexts, identifying their common frameworks, recent advances, and persistent limitations. This review reveals that such discrepancies primarily stem from inherent limitations of LLMs and flaws in simulation design, both of which are examined in detail. Building on these insights, we propose a systematic solution framework that emphasizes enriching data foundations, advancing LLM capabilities, and ensuring robust simulation design to enhance reliability. Finally, we introduce a structured algorithm that operationalizes the proposed framework, aiming to guide credible and human-aligned LLM-based simulations. To facilitate further research, we provide a curated list of related literature and resources at \url{https://github.com/Persdre/awesome-llm-human-simulation}.
\end{abstract}


\begin{figure}[h!]
    \centering
    \includegraphics[width=\linewidth]{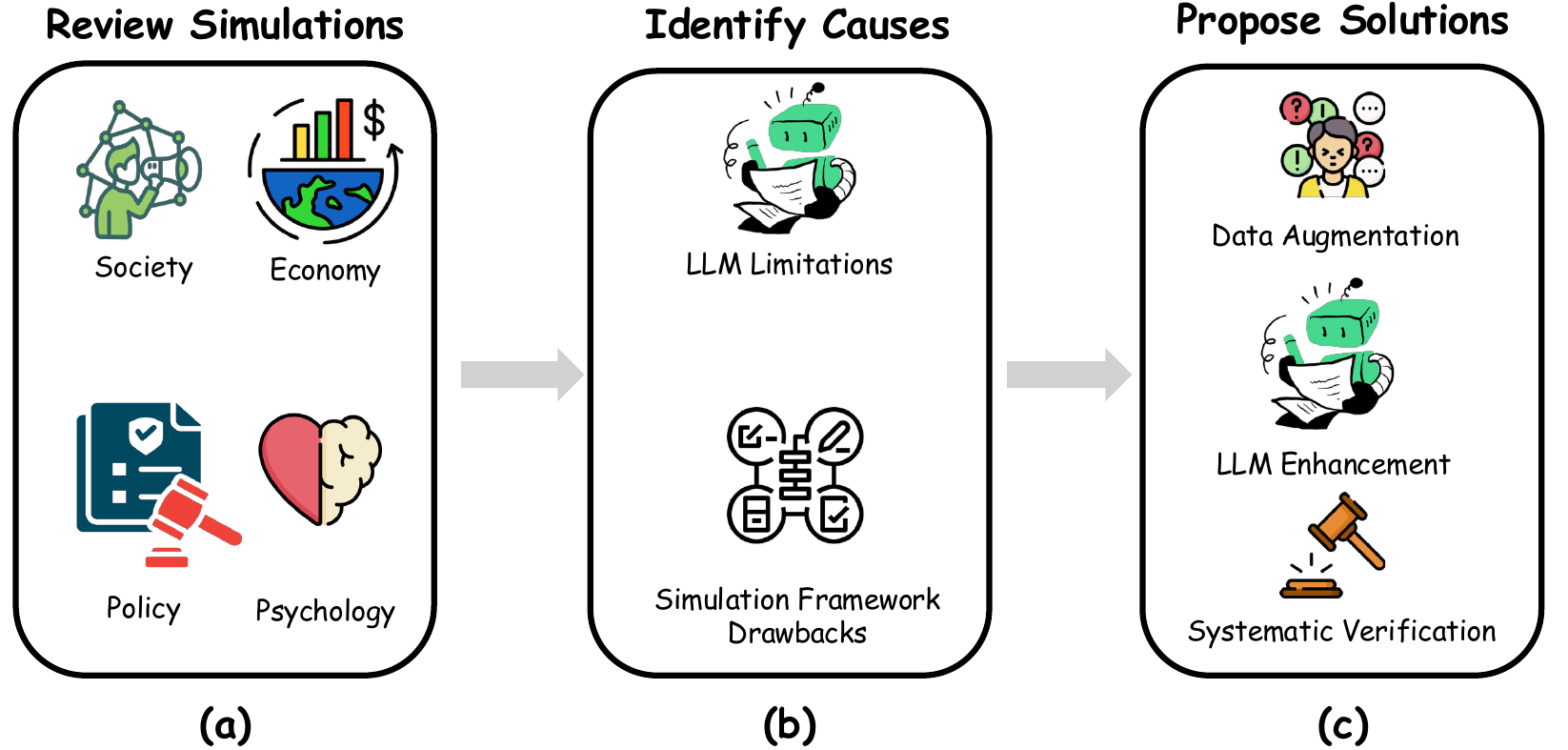}
    \caption{Flow of This Position Paper. We start by reviewing the current LLM-based human simulations, and then identify the causes of the gaps between simulation outputs and real-world human behavior. Finally, we propose targeted solutions for advancing the reliability of LLM-based human simulation.}
    \label{fig:overview}
\end{figure}
\vspace{-1em}


\section{Introduction}
Simulation has long been instrumental in understanding and replicating human actions and characteristics, driving progress in fields like social dynamics, policy analysis, and economic forecasting \cite{winsberg2003simulated, ofoegbu2023simulation, cioffi2010methodology, dilaver2023unpacking, orcutt1976policy}. The advent of Large Language Models (LLMs) has brought in a new era, with researchers leveraging LLM agents as proxies for humans within simulated environments \cite{Park2023GenerativeAgents, lin2023agentsims, li2023you, wu2024shall, zhang2024llm, shi2024learning, wang2024megaagent}. Initial successes span diverse domains, including societal modeling, economics, policy simulation, and psychology \cite{Park2023GenerativeAgents, lin2023agentsims, li2024cryptotrade, chen2024agentcourt, li2024agent, yang2024psychogat}. Furthermore, high-fidelity LLM simulations hold promise for generating valuable training data \cite{tang2024synthesizing, zhang2024regurgitative} and evaluating data quality \cite{xu2023reasoninglargelanguagemodels,moniri2024evaluatingperformancelargelanguage}, thus acting as both data generators and evaluators to enhance LLM capabilities \cite{gu2024survey, son2024llm, li2024salad}.

Despite these advancements, a critical question occurs: \textit{\textbf{Can current LLM-based simulations simulate humans reliably?}}

Our position is: \textbf{Current LLM-based human simulations are not reliable.} This position is supported by empirical evidence revealing significant discrepancies between LLM-generated simulations and observed human behavior. As summarized in Table~\ref{table:simulation_failures}, these shortcomings are consistently demonstrated by LLMs across diverse domains when tasked with simulating human actions and responses.

\begin{table*}[!ht]
    \centering
    \caption{Empirical evidence of gaps between LLM simulations and real human behavior.}
    \label{table:simulation_failures}
    \resizebox{\linewidth}{!}{%
    \begin{tabular}{c|c|c}
    \toprule
    \rowcolor{champagne}\textbf{Simulation Task} & \textbf{Metric} & \textbf{How LLM Performance Diverged from Human Behavior} \\
    \midrule
    \rowcolor{gray!10}Clinics Diagnosis \cite{schmidgall2024agentclinic} & Accuracy & LLM achieved only 40.18\% diagnostic correctness, significantly below human expert levels. \\
    Emotion Understanding \cite{li2024quantifying} & Accuracy & GPT-4 (58\%) demonstrated lower accuracy in understanding emotions compared to humans (70\%). \\
    \rowcolor{gray!10}Persona \cite{hu2024quantifying} & Variance & Assigned persona accounted for <10\% of variance in human annotations, indicating weak role adherence. \\
    Human-like Qualities \cite{wu2024self} & Accuracy & \makecell[l]{GPT-4's human-likeness rating (51\% at 3 turns) sharply \\ degraded over longer interactions (e.g., 13.3\% at 20 turns).} \\
    \rowcolor{gray!10}Trading \cite{li2024cryptotrade} & Return & GPT-4's trading return (28\%) significantly underperformed that of human traders (39\%). \\
    Adversarial Attack \cite{chen2024humans} & Attack Success Rate & GPT-4-Turbo (27\% success rate) proved far more susceptible to attacks than humans (6\%). \\
    \bottomrule
    \end{tabular}%
    }
\end{table*}

By conducting a thorough review of current LLM-based human simulations across social, economic, policy, and psychological domains, we attribute the gaps to two main categories:

\textbf{(1) The inherent limitations of current LLMs} in capturing the nuances of human cognition and behavior. This first issue manifests through multiple dimensions: (a) embedded biases (cultural, gender, occupational, and socioeconomic) that distort behavioral simulations \cite{kotek2023gender, wan2023kelly, wang2023not}; (b) cognitive process limitations that compromise decision-making authenticity \cite{li2024quantifying, gui2023challenge}; (c) inconsistent behavioral patterns over time due to memory constraints \cite{zheng2024towards,zhong2024memorybank}; and (d) interaction mechanism deficiencies that affect multi-agent simulation quality \cite{tang2025the, li2024hello}.

\textbf{(2) The design flaws of current simulation frameworks} in adequately modeling real-world complexities and interactions. These flaws include: (a) oversimplification of complex psychological states and interactions \cite{williams2000emotion}; (b) insufficient incorporation of comprehensive human experiences and contexts \cite{grossberg1987neural,chen2024humans}; (c) inadequate modeling of human incentives and motivations \cite{stern1999information, petrakis2012human}; (d) limited validation mechanisms lacking comprehensive evaluation criteria \cite{he2023towards, ke2024exploring}; and (e) inadequate real-time monitoring and adjustment mechanisms \cite{xexeo2024economic, reason2024artificial}. 

To address the above gaps, we propose a systematic solution framework composed of three parts. Firstly, we advocate for enriching the foundational data used in simulations to ensure more realistic and diverse inputs. Secondly, to improve LLM-based agents' abilities in simulation, we outline targeted interventions for mitigating inherent LLM limitations—focusing on their cognitive process simulation, behavioral consistency, and memory capabilities. Lastly, we propose principles for verifiable and robust simulation frameworks, emphasizing continuous and multi-level validation architectures. Finally, we culminate this framework in Algorithm~\ref{algo:solution}, an operational algorithm that synthesizes these targeted solutions into a structured workflow for developing, executing, and validating reliable LLM-based human simulations.

Our Contributions are summarized as follows:
\begin{itemize}[leftmargin=*, itemsep=0pt, topsep=2pt] 
    \item We conduct a systematic analysis of LLM-based human simulations, pinpointing inherent LLM limitations and simulation framework design flaws as the two central challenges hindering simulation fidelity across various applications.
    \item We propose targeted solutions across three key areas: enhancing the data foundations for simulation, mitigating LLM limitations in cognitive and behavioral modeling, and establishing methodologies for verifiable and robust simulation framework design.
    \item We propose a structured algorithm that synthesizes these targeted solutions into a structured workflow for developing, executing, and validating reliable LLM-based human simulations.
\end{itemize}

\section{Current LLM-based Human Simulations} \label{sec:review}

LLM-based human simulations have shown initial success across diverse fields. Fundamentally, these simulations revolve around two core dimensions: \textbf{LLM-driven Capabilities} and the overarching \textbf{Simulation Design}. This section begins by presenting a general formulation for LLM-based human simulations, outlining their foundational components. Following this, we will explore four primary categories of these simulations: social, economic, policy and psychological. For each category, we will examine its applications, effectiveness and limitations, with a specific analysis of how LLM-driven capabilities are utilized and how the simulation framework is structured.

\subsection{LLM-based Human Simulation Formulation} \label{sec:formation}

We formalize LLM-based human simulation as a tuple
\begin{equation}
\mathcal{H} = (\mathcal{E}, \mathcal{F}, \mathcal{R}), 
\tag{1}
\end{equation}
where the environment $\mathcal{E} = (\mathcal{S}, \mathcal{V})$ includes the state space $\mathcal{S}$ and evaluation procedures $\mathcal{V}$, the agents $\mathcal{F} = \{ f_1, \dots, f_n \}$ are LLM-based policies, and the rules $\mathcal{R}$ define interaction mechanisms, state transitions, and evaluation criteria.

Each agent $f_i$ maps its input message $m$ and current state $s \in \mathcal{S}$ to an action $a$: 
\begin{equation}
f_i: (m, s) \mapsto a. 
\tag{2}
\end{equation}

LLMs provide the \textbf{LLM-driven Capabilities} enabling $f_i$ to generate human-like behaviors, while empirical data and domain knowledge ground $\mathcal{V}$ and inform the overall \textbf{Simulation Design}.

\subsection{Social Simulation}

Within the formalism $\mathcal{H} = (\mathcal{E}, \mathcal{F}, \mathcal{R})$, social simulations define 
the environment $\mathcal{E}$ as a communicative and relational space, where $\mathcal{S}$ encodes evolving conversational and social states, and $\mathcal{V}$ evaluates dialogue plausibility, role consistency, and emergent group dynamics. 
Agents $f_i \in \mathcal{F}$ map $(m, s) \mapsto a$ into conversational actions, while $\mathcal{R}$ governs turn-taking, interactional norms, and social coordination rules.

\noindent \textbf{LLM-driven Capabilities.}  
LLMs enable agents to generate realistic dialogue, make nuanced social decisions, and embody distinct personas. These capabilities allow emergent social phenomena to be explored under controlled but flexible settings.

\noindent \textbf{Simulation Design.}  
Framework design focuses on specifying $\mathcal{E}$ (communication protocols, social settings) and $\mathcal{R}$ (interaction rules, persona constraints), with $\mathcal{V}$ incorporating validation metrics such as coherence, plausibility, and cross-agent consistency. Major challenges include validating long-horizon behaviors and capturing authentic emergent processes.

\noindent \textbf{Representative Implementations.}  
Applications include virtual towns for daily life modeling \cite{park2023generative}, inter-agent competition \cite{zhao2023competeai}, healthcare simulations (Agent Hospital \cite{li2024agent}, AgentClinic \cite{schmidgall2024agentclinic}), and peer review systems \cite{jin2024agentreview}. These showcase the adaptability of LLM-based social simulations.

\noindent Despite their versatile applications, significant limitations persist. We summarize the limitations in social simulation as follows:

\begin{tcolorbox}[
  colback=teal!15,                
  colframe=teal!60!black,         
  boxrule=0.8pt,
  arc=2mm,
  left=5pt, right=5pt, top=5pt, bottom=5pt, 
  title=\textbf{Limitations}
]
\noindent (1) \textbf{Persona Inconsistency:} Agents struggle to maintain stable, long-term personality traits throughout interactions. (2) \textbf{Trait-Behavior Mismatch:} A significant gap exists between agents' self-reported characteristics (e.g., ``extraverted") and their actual observed ``introverted" behaviors. (3) \textbf{Fragile Social Dynamics:} The coherence of multi-agent social systems is often brittle, facing challenges in sustaining believable long-horizon dynamics.
\end{tcolorbox}


\subsection{Economic Simulation}

In economic simulations, $\mathcal{E}$ specifies market environments, with $\mathcal{S}$ representing market states (e.g., order books, prices) and $\mathcal{V}$ providing metrics such as efficiency, returns, and stability. 
Agents $f_i \in \mathcal{F}$ implement policies $(m, s) \mapsto a$ as economic actions like bidding, trading, or resource allocation, governed by rules $\mathcal{R}$ encoding market protocols and strategic constraints.

\noindent \textbf{LLM-driven Capabilities.}  
LLMs provide adaptive decision-making for trading, portfolio management, and negotiation. They exhibit limited rationality, time preferences, and occasionally fairness-driven or cooperative behaviors resembling human economic agents.

\noindent \textbf{Simulation Design.}  
Design choices define $\mathcal{E}$ as the market system, $\mathcal{R}$ as trading rules, and $\mathcal{V}$ as validation against benchmarks or theoretical equilibria. Frameworks must ensure market stability, realism of agent preferences, and robustness of evaluation mechanisms.

\noindent \textbf{Representative Implementations.}  
Examples include CryptoTrade for cryptocurrency markets \cite{li2024cryptotrade}, EconAgent for labor and macroeconomic simulations \cite{li2024econagent}, and agent-based analyses of Nash equilibria and cooperation \cite{guo2024economics, fontana2024nicer}. Studies report both alignment with utility-maximizing principles and deviations reflecting fairness or irrationality \cite{bybee2023surveying, ross2024llm}.

\noindent However, despite capturing some human-like deviations, crucial differences in economic behavior persist. We summarize the primary limitations below:

\begin{tcolorbox}[
  colback=teal!15, colframe=teal!60!black,
  boxrule=0.8pt, arc=2mm, left=5pt, right=5pt, top=5pt, bottom=5pt,
  title=\textbf{Limitations}
]
\noindent (1) \textbf{Weaker Loss Aversion:} Compared to humans, LLM agents exhibit a diminished aversion to losses, affecting their risk-taking behavior. (2) \textbf{Stronger Time Discounting:} Agents devalue future rewards more heavily than humans, showing a stronger preference for immediate gratification. (3) \textbf{Limited Fidelity:} These behavioral gaps limit the fidelity of simulations in accurately capturing human economic decision-making in certain contexts.
\end{tcolorbox}


\subsection{Policy Simulation}

For policy simulations, $\mathcal{E}$ encodes institutional environments with $\mathcal{S}$ representing societal states (e.g., economic indicators, compliance levels) and $\mathcal{V}$ measuring impacts such as equity or efficiency. 
Agents $f_i \in \mathcal{F}$ model stakeholders whose actions $(m, s) \mapsto a$ reflect policy responses, while $\mathcal{R}$ defines implementation mechanisms, enforcement, and adaptation rules.

\noindent \textbf{LLM-driven Capabilities.}  
LLMs generate responses to policy interventions, simulate compliance vs. resistance, and capture diverse stakeholder perspectives. They can provide foresight into short-term responses and interactional dynamics.

\noindent \textbf{Simulation Design.}  
Design involves specifying $\mathcal{E}$ as the policy context, $\mathcal{R}$ as regulatory and enforcement structures, and $\mathcal{V}$ as outcome assessment metrics (e.g., societal welfare, inequality, public trust). Challenges lie in representing heterogeneous stakeholder interests and cascading long-term impacts.

\noindent \textbf{Representative Implementations.}  
Notable systems include EconAgent for economic policy analysis \cite{li2024econagent} and Urban Generative Intelligence (UGI) frameworks with domain-specific LLMs for urban planning \cite{xu2023urban, zou2025deep}. These applications show the potential for simulating systemic policy dynamics.

\noindent Despite this potential for simulating systemic dynamics, these frameworks face significant hurdles. The most pressing limitations are summarized as follows:

\begin{tcolorbox}[
  colback=teal!15, colframe=teal!60!black,
  boxrule=0.8pt, arc=2mm, left=5pt, right=5pt, top=5pt, bottom=5pt,
  title=\textbf{Limitations}
]
\noindent (1) \textbf{Modeling Long-Term Effects:} Accurately capturing the cascading, long-term impacts of policies and the complex interplay of diverse stakeholder interests remains a formidable challenge. (2) \textbf{Inherent LLM Biases:} Biases within the models can produce skewed policy interpretations and inequitable simulated outcomes if not carefully mitigated. (3) \textbf{Validation Challenges:} The multifaceted impacts predicted by simulations are difficult to validate due to the immense cost and logistical complexity of large-scale, real-world verification.
\end{tcolorbox}


\subsection{Psychological Simulation}

In psychological simulations, $\mathcal{E}$ specifies experimental contexts, with $\mathcal{S}$ encoding cognitive and affective states, and $\mathcal{V}$ assessing validity against psychological theories or empirical data. 
Agents $f_i \in \mathcal{F}$ map $(m, s) \mapsto a$ into cognitive or behavioral outputs, while $\mathcal{R}$ defines task protocols, persona constraints, and state transitions.

\noindent \textbf{LLM-driven Capabilities.}  
LLMs can emulate cognitive processes such as decision reasoning, mental state inference, and trait expression. They generate behavioral responses reflecting psychological constructs and dynamic mental-state changes.

\noindent \textbf{Simulation Design.}  
Framework design encodes experimental protocols, role assignments, and evaluation criteria. $\mathcal{V}$ typically grounds evaluation in psychological validity, response consistency, and alignment with established findings.

\noindent \textbf{Representative Implementations.}  
Applications include ChatCounselor for counseling interactions \cite{liu2023chatcounselor}, PsychoGAT for trait inference \cite{yang2024psychogat}, and PATIENT-$\psi$ for medical training \cite{wang2024patient}. Studies also benchmark LLMs in cognitive tasks, showing near-human performance in specific reasoning tests \cite{binz2023using, hagendorff2023machine}.

\noindent While these applications show promise in mimicking specific cognitive tasks, a core challenge constrains their psychological fidelity. This primary limitation is outlined below:

\begin{tcolorbox}[
  colback=teal!15, colframe=teal!60!black,
  boxrule=0.8pt, arc=2mm, left=5pt, right=5pt, top=5pt, bottom=5pt,
  title=\textbf{Limitations}
]
\noindent The primary challenge is a \textbf{core representational difficulty}: LLMs struggle to authentically model and represent underlying, long-term psychological traits. This fundamental issue constrains the fidelity and depth of simulations across various applications, from counseling to cognitive assessment, even as they show success in reproducing specific behaviors.
\end{tcolorbox}


\section{LLM Inherent Drawbacks}
\label{sec:limitations}
Drawing from the limitations highlighted in Section \ref{sec:review}, we now analyze the intrinsic properties of LLMs that fundamentally constrain their use as reliable simulators.

\subsection{Bias}

\noindent \textbf{Bias in LLMs fundamentally affects cognitive and behavioral simulation.} Experiments in utility theory \cite{ross2024llm} show that only two out of nine models passed all competence tests compared with human responses, with stronger time discounting, less rational loss assessments, and greater inequity aversion. Such distortions directly limit the authenticity of simulated reasoning and decision-making.

\noindent \textbf{Cultural bias undermines cross-cultural reliability.} Because training data are predominantly English and Western, LLMs systematically favor Western values and interaction patterns. Comparative studies report significantly higher accuracy in U.S. contexts compared to non-Western settings such as Japan and South Africa \cite{wang2023not, qu2024performance}, highlighting the lack of generalizable cultural cognition.

\noindent \textbf{Gender bias distorts behavioral pattern simulation.} LLMs perpetuate stereotypes at far higher rates than humans, with models generating 3–6 times more gender-stereotypical behaviors. Alignment efforts remain insufficient: GPT-4-Turbo shows a 27\% attack bias success rate compared to 6\% in humans \cite{kotek2023gender, chen2024humans}.

\noindent \textbf{Socioeconomic and occupational bias reduces diversity.} Overrepresentation of certain professions and social groups in training data leads to unrealistic reasoning about underrepresented groups \cite{gwartney1971variance, harrison2015unintended}, undermining the ability to model authentic diversity in human decision-making.

\subsection{Mismatches in Simulating Cognition and Behavior}

\noindent \textbf{Cognitive mismatches reduce decision-making authenticity.} LLMs display inconsistent reasoning patterns across tests. For instance, Mixtral-8*7b scores low extraversion (2) in one test but high (5) in another, while value surveys drop from 90\% to 70\% under adversarial prompts \cite{li2024quantifying, hu2024quantifying}. These mismatches indicate fragile reasoning, weak emotional processing, and limited adaptability. 

A further complication arises from \emph{experiment-awareness}. In human-subject research, participants are typically kept unaware of the precise research purpose to avoid demand characteristics. In contrast, LLMs may recognize that they are being tested and adapt their outputs accordingly. Recent work shows that frontier models can detect evaluation settings and even infer the intended task, leading to altered behaviors. \cite{needham2025evalaware, nguyen2025probing}.


\noindent \textbf{Behavioral patterns lack temporal consistency.} Studies in generative simulation (e.g., GovSIM) show that models fail to sustain equilibrium, leading to unsustainable resource use such as overharvesting \cite{piatti2024cooperate}. Similarly, LLMs fail to form realistic habits \cite{zhong2024memorybank} or simulate learning over repeated interactions \cite{chu2024improve}.


\noindent \textbf{Data contamination drives memorization.} Because training corpora overlap with benchmarks, LLMs sometimes recall memorized content instead of reasoning. For example, MMLU subsets contain up to 57\% flawed items \cite{gema2024mmlu}, and GPT-4 can predict masked options with 57\% exact-match accuracy \cite{deng2024investigating}. This undermines the authenticity of “simulated reasoning.”

\noindent \textbf{Memory constraints limit behavioral consistency across interactions.} LLMs' memory limitations particularly affect the behavioral patterns and interaction mechanisms components \cite{tang2025the}. As cognitive load increases, LLMs demonstrate severe performance degradation; for instance, at the highest load levels, performance can decline by 50\% to 60\% relative to initial levels observed at low cognitive loads \cite{upadhayay2024cognitive}. These constraints significantly impact their ability to model long-term relationships \cite{li2024hello}, retain context across multiple interactions \cite{yuan2024back}, and adapt behaviors based on past experiences \cite{evans2015and}.

\section{Simulation Design Drawbacks}
\label{sec:simulation_drawbacks}
\noindent Human-designed simulation frameworks also face critical limitations that undermine their reliability as analyzed in Section \ref{sec:review}. We summarizes these drawbacks in Table~\ref{tab:sim_design_drawbacks}, while the following subsections elaborate with representative cases. 

Overall, these limitations can be grouped into two major categories. The first set of issues arises from the \colorbox{lightpurple}{\textbf{design of simulation frameworks themselves}}, 
where human cognition, experience, and incentives are often simplified or incompletely modeled. 
The second set concerns \colorbox{lightpurple}{\textbf{validation and monitoring}}, where the lack of rigorous evaluation and adaptive mechanisms decrease reliability of simulations. 
The following subsections discuss these two aspects in detail.

\begin{table*}[!ht]
\centering
\resizebox{0.9\linewidth}{!}{%
\begin{tabular}{p{3.5cm}|p{5cm}|p{7cm}}
\toprule
\textbf{Drawback Type} & \textbf{Impact on Simulation} & \textbf{Representative Evidence} \\
\midrule
\rowcolor{lightpurple}Oversimplified psychological states & Reduces realism of cognition and group dynamics & Emotional states collapsed into basic categories or scales; group dynamics poorly captured \cite{tjuatja2024llms, jansen2023employing, williams2000emotion} \\
Incomplete coverage of lived experiences & Neglects historical, cultural, and contextual richness in decisions & CRS simulations degrade after 2–5 rounds due to lack of experiential integration \cite{beratan2007cognition, zhu2024reliable} \\
\rowcolor{lightpurple}Weak incentive modeling & Fails to capture complex social and cultural motivations & Simplified rewards ignore ``face'' or reputation trade-offs; unrealistic decision-making \cite{stern1999information, hwang1987face} \\
\midrule
Validation gaps & Lacks comprehensive evaluation metrics for behaviors and interactions & Metrics insufficient to capture complex dynamics; authenticity criteria unclear \cite{he2023towards, ke2024exploring} \\
\rowcolor{lightpurple}Limited monitoring & Poor adaptability to emergent or unexpected behaviors & Current systems lack real-time adjustment; unable to maintain simulation quality \cite{xexeo2024economic, reason2024artificial} \\
\bottomrule
\end{tabular}%
}
\caption{Summary of simulation design drawbacks that limit reliability and authenticity in LLM-based human simulations.}
\label{tab:sim_design_drawbacks}
\end{table*}

\definecolor{lightpurple}{RGB}{235,223,249} 

\subsection{Framework Design Drawbacks}

\noindent \textbf{Current frameworks oversimplify complex human psychological states and interactions.} Human designers often create oversimplified frameworks when attempting to model psychological states \cite{tjuatja2024llms, jansen2023employing}. For example, many frameworks reduce complex emotional states to basic categories or numerical scales, failing to capture the subtle interplay between different psychological factors \cite{williams2000emotion}. This oversimplification stems from the challenge designers face in quantifying and operationalizing complex human psychological processes \cite{evans2015and}. In addition, while LLMs have shown promise in simulating individual human behavior, simulating group behaviors introduces additional complexities. Group dynamics are influenced by factors such as social norms, interpersonal relationships, and collective decision-making processes, which are not easily captured by current LLM frameworks.

\noindent \textbf{Simulation designs fail to account for comprehensive human experiences.} Framework designers struggle to incorporate the full spectrum of human lived experiences into their simulations \cite{beratan2007cognition}. Current designs often focus on specific, measurable behaviors while neglecting the rich tapestry of personal histories, cultural contexts, and life experiences that shape human decision-making \cite{grossberg1987neural,chen2024humans}. For example, in conversational recommender system simulations, the success rate of recommendations decreased significantly under 2-5 rounds (from 70\% to only 1\%) with the user simulator indicating the challenge of CRS not effectively utilizing the interaction information obtained from user simulators, due to the complexity of scenarios \cite{zhu2024reliable}. This limitation reflects the difficulty in creating frameworks that can adequately represent the complexity of human experiential learning.

\noindent \textbf{Current frameworks lack effective human incentive modeling.} Designers face significant challenges in creating frameworks that accurately model complex human motivations and incentives \cite{stern1999information, petrakis2012human}. Many current designs rely on simplified reward systems that fail to capture the intricate web of personal, social, and cultural factors influencing human decision-making. For instance, individuals often make decisions contrary to their personal preferences to maintain social status, reputation, or 'face' in certain cultural contexts \cite{hwang1987face}. This limitation stems partly from LLMs' inherent constraints - they lack embodied experience and real social interactions, hindering their ability to fully comprehend complex social dynamics. Such oversimplification results in behavioral simulations that inadequately reflect the true complexity of human motivational systems.

\subsection{Validation and Monitoring Drawbacks}

\noindent \textbf{Current validation mechanisms lack comprehensive evaluation criteria.} Human-designed validation systems struggle to establish effective criteria for evaluating simulation authenticity \cite{he2023towards}. The challenge lies in developing metrics that can effectively measure both the accuracy of individual behaviors and the coherence of complex interaction patterns \cite{ke2024exploring}. Current frameworks often rely on oversimplified validation methods that fail to capture the full complexity of human behavior.

\noindent \textbf{Monitoring systems lack effective real-time adjustment capabilities.} Framework designers struggle to create effective mechanisms for real-time monitoring and adjustment of simulation parameters \cite{xexeo2024economic}. Current designs often lack the flexibility to adapt to emerging behavioral patterns or unexpected interaction dynamics, limiting their ability to maintain simulation quality \cite{reason2024artificial}.

\section{Proposed Solutions}
\label{sec:solutions}
\begin{figure}[t]
    \centering
    \includegraphics[width=0.48\textwidth]{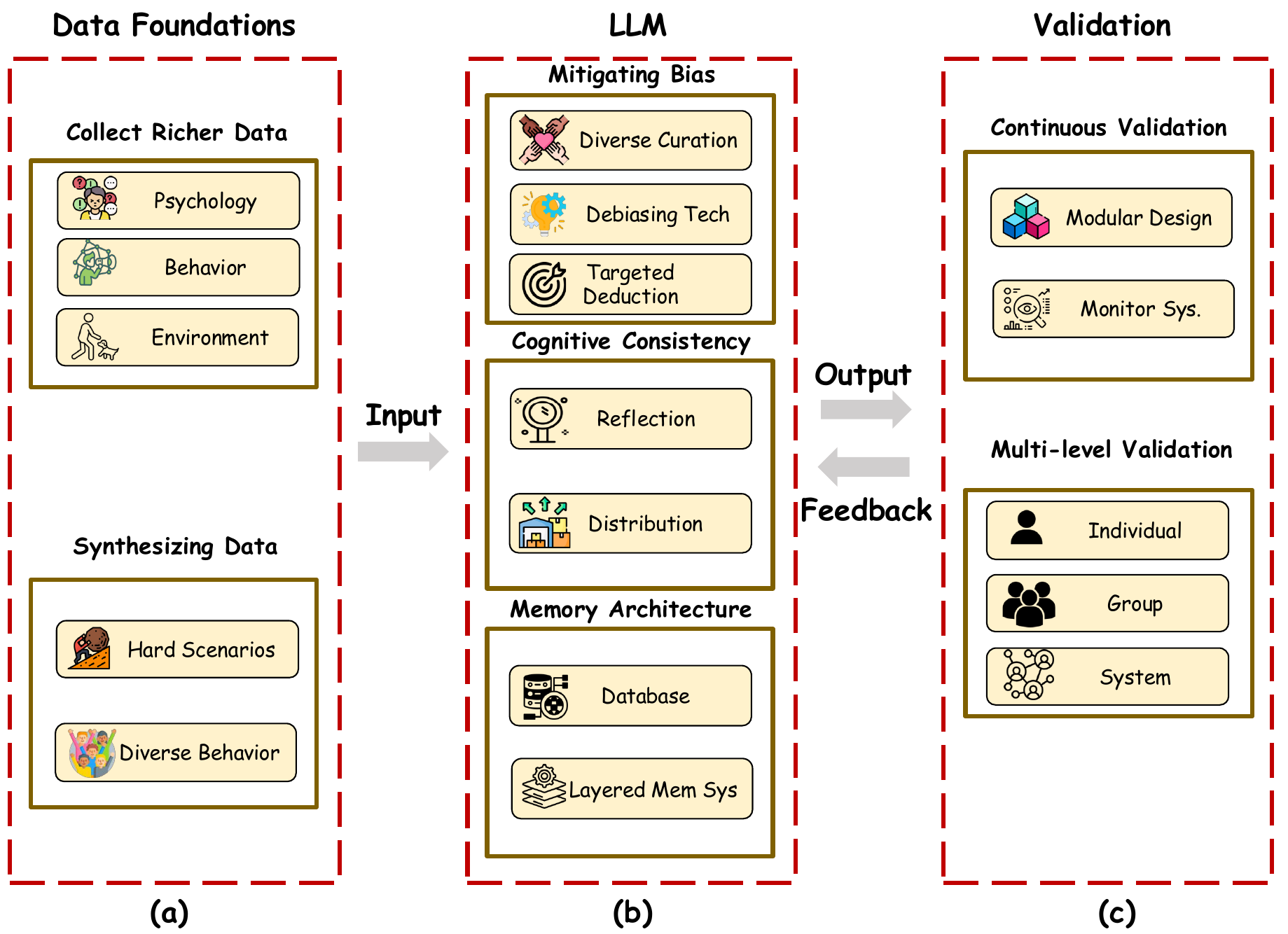}
    \caption{Overview of the Proposed Solution Framework. It details three core components: (a) Enriched Data Foundations, (b) Improved LLM Capabilities, and (c) Trustworthy Simulation Design through Robust Validation.}
    \label{fig:solutions}
\end{figure}

We propose systematic solutions that directly address the limitations identified in Sections \ref{sec:limitations} and \ref{sec:simulation_drawbacks}. We focus on three areas: (1) enriching the data foundation (Section~\ref{sec:sol-data}), (2) improving the abilities of LLM (Section~\ref{sec:sol-llm}), and (3) engineering trustworthy simulations through robust validation (Section~\ref{sec:simulation-design}). Figure~\ref{fig:solutions} provides an overview.


\begin{algorithm*}[t]
\caption{Our Proposed LLM-based Human Simulation Framework}
\label{algo:solution}

\small 

\begin{algorithmic}[1]
\STATE \textbf{Initialize:} Agent set $\mathcal{F} = \{f_1, ..., f_n\}$, Environment $\mathcal{E}$, Validation metrics $\mathcal{V}$, History $H \leftarrow \emptyset$ \hfill \textcolor{gray}{// Setup simulation components}

\WHILE{simulation not complete}
    \FOR{each agent $f_i \in \mathcal{F}$}
        \STATE Perception update: $p_i \leftarrow \text{perceive}(f_i, \mathcal{E}, H)$ \hfill \textcolor{gray}{// Agent observes environment}
        \STATE Generate action: $a_i \leftarrow f_i(p_i, \mathcal{E})$ \hfill \textcolor{gray}{// Context-aware action generation}
        \STATE Behavioral validation: $v_i \leftarrow \mathcal{V}_{\text{behav}}(a_i, D_{\text{human}})$ \hfill \textcolor{gray}{// Compare with human patterns}
        \IF{$v_i < \text{threshold}$}
            \STATE $a_i \leftarrow \text{adjust}(a_i, D_{\text{human}})$ \hfill \textcolor{gray}{// Recalibrate non-human-like actions}
        \ENDIF
        \STATE Execute action: $\mathcal{E} \leftarrow \text{apply}(a_i, \mathcal{E})$ \hfill \textcolor{gray}{// Apply action to environment}
        \STATE Memory update: $M_i \leftarrow M_i \cup \{(p_i, a_i, \mathcal{E})\}$ \hfill \textcolor{gray}{// Update agent memory}
    \ENDFOR
    
    \STATE 
    \IF{$\text{time} \bmod \text{validate\_interval} == 0$}
        \STATE Individual validation: $V_{\text{indiv}} \leftarrow \mathcal{V}_{\text{indiv}}(\{a_i\}, D_{\text{human}})$ \hfill \textcolor{gray}{// Assess individual behaviors}
        \STATE Group validation: $V_{\text{group}} \leftarrow \mathcal{V}_{\text{group}}(\mathcal{E}, D_{\text{social}})$ \hfill \textcolor{gray}{// Verify group dynamics}
        \STATE Expert assessment: $V_{\text{expert}} \leftarrow \mathcal{V}_{\text{expert}}(\mathcal{E}, H)$ \hfill \textcolor{gray}{// Optional human expert review}
        
        \STATE Apply calibrations: $\mathcal{F} \leftarrow \text{calibrate}(\mathcal{F}, [V_{\text{indiv}}, V_{\text{group}}, V_{\text{expert}}])$ \hfill \textcolor{gray}{// Adjust agent parameters}
    \ENDIF
    
    \STATE Anomaly detection: $A \leftarrow \text{detect\_anomalies}(\mathcal{E}, H, \mathcal{F})$ \hfill \textcolor{gray}{// Identify unrealistic patterns}
    \STATE Environment correction: $\mathcal{E} \leftarrow \text{correct}(\mathcal{E}, A)$ \hfill \textcolor{gray}{// Fix detected anomalies}
    \STATE Update history: $H \leftarrow H \cup \{(\{a_i\}, \mathcal{E}, V)\}$ \hfill \textcolor{gray}{// Record simulation state}
\ENDWHILE

\STATE \textbf{Return:} Validated simulation results $\mathcal{E}$, Interaction history $H$, Validation metrics $V$
\end{algorithmic}
\end{algorithm*}



\subsection{Enriching Data Foundation for Simulations}\label{sec:sol-data}
\subsubsection{Harnessing Richer Human Data} 
Simulations grounded solely in traditional text, image, or audio data inevitably miss crucial dimensions of human experience.

\noindent \textbf{Capturing Physiological and Cognitive Data.} Wearable sensors unlock access to vital physiological signals, including heart rate and skin conductance, and can even record neural activity patterns, offering deeper insights into internal states \cite{yuan2024back}. Integrating such data is key for LLMs to move beyond superficial behavior mimicry towards modeling the underlying emotional states and cognitive processes that drive human decisions.

\noindent \textbf{Monitoring Fine-Grained Behavioral Patterns.} Motion sensing and activity tracking technologies provide continuous, detailed records of daily routines, physical movement, and social interaction dynamics, as demonstrated in studies like \cite{chu2024improve}. This granular behavioral data enables LLMs to generate actions that are not just plausible in general, but accurate within specific situational and personal contexts.

\noindent \textbf{Achieving Environmental Context Awareness.} Human behavior is deeply embedded in its environmental context. Incorporating data from sensors that measure ambient conditions—such as light, sound, and temperature \cite{li2024hello}—allows simulations to account for these crucial external factors, leading to more nuanced and realistic models of human responsiveness.

\subsubsection{Synthesizing High-Fidelity Human Data via LLMs}
While real-world data is invaluable, its collection is often constrained by cost, logistical complexity, ethical barriers, and the sheer impossibility of observing rare or counterfactual scenarios. LLMs offer a powerful complementary approach by enabling the synthesis of high-quality data to augment or replace real-world data where necessary.

\noindent \textbf{Generating Data for Inaccessible Scenarios.} LLMs excel at creating plausible data for situations that are difficult, dangerous, or unethical to capture in reality. Examples include simulating high-stakes negotiations, sensitive social interactions, or specific therapeutic encounters \cite{tjuatja2024llms, yang2024social}. This capability provides crucial training and evaluation data that would otherwise be unavailable.

\noindent \textbf{Synthesizing Diverse Behaviors.} Human behavior is inherently diverse. To effectively model this, LLMs can generate a wide spectrum of behavioral patterns through the manipulation of key parameters. These parameters can include intrinsic agent characteristics such as personality traits, emotional states, and cultural backgrounds \cite{he2023towards}. Furthermore, LLMs can synthesize varied behavioral responses under specific, but infrequently occurring conditions. For instance, they can be used to augment datasets by generating human behaviors in bull or bear markets, allowing for the exploration of human reactions within these distinct economic markets \cite{li2024cryptotrade, Guo2024EconomicsAF}. 

\subsection{Improving LLM Abilities} \label{sec:sol-llm}
The persistence of cultural, gender, and demographic biases, even in state-of-the-art LLMs \cite{jiao2024navigating, foka2024she}, necessitates proactive mitigation strategies that go beyond standard training procedures. We propose a multi-faceted approach to tackle these challenges as follows:

\noindent \textbf{Mitigating Bias through Advanced Training.} Building on recent advances in bias understanding and mitigation \cite{pawar2024survey, rakshit2024prejudice}, we advocate for a multi-pronged framework. This involves: (1) Diverse Data Curation, by actively incorporating balanced, multilingual datasets reflecting diverse cultural and demographic groups to counteract learned biases; (2) Algorithmic Debiasing, implementing techniques like counterfactual data augmentation and balanced training objectives during model development to reduce learned biases; and (3) Targeted Bias Reduction, such as occupation-aware post-training to prevent specific stereotype perpetuation.

\noindent \textbf{Architectures for Believable and Consistent Cognition.} To improve the believability and consistency of simulated human cognition, we propose architectural innovations. These include: (1) Internal Reflection Mechanisms, enabling agents to internally review and align planned responses with their persona, goals, and past actions before outputting \cite{li2024quantifying}. (2) {Distributed Cognitive Architecture using specialized agents for distinct cognitive modules (e.g., planning, reasoning, emotion). This modular approach, unlike monolithic models, improves performance and consistency in specific cognitive domains, as supported by psychology and economic simulations \cite{hu2024quantifying, wang2024enhancing}.

\noindent \textbf{Overcoming Memory Deficits with Hybrid Architectures.} Naive approaches to augmenting LLMs with external memory often fall short of resolving fundamental limitations like constrained context windows, retrieval inaccuracies, and information forgetting \cite{wang2023augmenting, hatalis2023memory, han2024review, feng2024far}. To address these limitations, we propose integrated hybrid memory systems. One key component, \textit{Scalable External Knowledge Bases}, utilizes vector stores or specialized databases for efficient large-scale information storage and precise retrieval, decoupling an agent's knowledge capacity from LLM constraints \cite{zhong2024memorybank}. Complementing this, \textit{Layered Memory Systems} with dynamic information management, such as short-term working and long-term episodic/semantic memory, akin to human cognitive structures \cite{yuan2024back, cowan2008differences} for more human-like recall and forgetting.

\subsection{Engineering Trustworthy via Validation} \label{sec:simulation-design}
Beyond improving LLMs themselves, the underlying simulation infrastructure must be engineered for reliability and robust validation. We propose a verification-centric approach with two key components:

\noindent \textbf{Modular Design with Continuous Validation.} Simulation frameworks should employ modular architectures where individual components, such as agent models or environmental modules, can be independently developed, tested, and validated \cite{reason2024artificial}. This should be paired with integrated monitoring systems that track agent behavior and system states to detect anomalies or deviations from expected patterns in real-time \cite{xexeo2024economic}. Modular design enables validation at multiple levels of granularity.

\noindent \textbf{Multi-level Validation Approach for Holistic Assessment.} Building on the above modular designs, a verifiable simulation framework implements validation at three critical levels for comprehensive assessment: (1) Individual-level validation: focusing on comparing individual agent responses and actions against statistical distributions derived from human behavior in similar contexts; (2) Group-level validation: assessing whether emergent social patterns and collective behaviors within the simulation match known human social phenomena or theoretical expectations of group dynamics; and (3) System-level validation: employing a combination of automated metrics and periodic expert review to ensure the holistic realism and plausibility of the simulation as a whole.


Synthesizing the solutions from the preceding subsections, we introduce Algorithm~\ref{algo:solution}, which, in accordance with the symbols defined in Section \ref{sec:formation}, provides a systematic framework for executing and validating LLM-based human simulations, thereby advancing their fidelity and trustworthiness.

\section{Rebuttal to Alternative Views}
\label{sec:discussion}

\begin{tcolorbox}[colback=yellow!15, colframe=yellow!50!black,
  boxrule=0.8pt, arc=2mm, left=4pt, right=4pt, top=4pt, bottom=4pt]
\textit{\textbf{Alternative View 1}: Functional Adequacy in Niche Applications.}
\end{tcolorbox}

\noindent \textbf{Position:} One perspective posits that LLM-based human simulations can achieve functional adequacy for particular applications, even without perfectly replicating human behavior \cite{park2024generative, yang2025twinmarket, wang2023roleplay}.

\noindent \textbf{\textit{Our Rebuttal:}} We contend that focusing on such "functional adequacy" is insufficient and potentially misleading. Our key counter-arguments are threefold: (1) \textbf{Brittleness over Robustness:} Success in isolated tasks is not a reliable indicator of general performance, as this perceived adequacy is often brittle and fails when conditions change even slightly. (2) \textbf{Lack of Generalization:} These successes fail to generalize; a model adequate for one scenario remains untrustworthy in broader applications without a proven foundation of reliability. (3) \textbf{The Need for Verifiability:} True progress demands moving beyond ad-hoc performance. The core aim of our work is to establish methodologies for building and verifying reliability, which is a prerequisite for any trustworthy application.

Beyond the pragmatic argument for functional adequacy, a more fundamental challenge addresses the very nature of human behavior itself, which is shown as follows:

\begin{tcolorbox}[colback=yellow!15, colframe=yellow!50!black,
  boxrule=0.8pt, arc=2mm, left=4pt, right=4pt, top=4pt, bottom=4pt]
\textit{\textbf{Alternative View 2}: The Absence of a Definitive Ground Truth in Human Behavior.}
\end{tcolorbox}

\noindent \textbf{Position:} This more fundamental view asserts that the quest for "reliable" simulation is inherently flawed, as human behavior itself lacks a singular, definitive ground truth. Proponents highlight the significant variance in human actions due to non-replicable contextual factors \cite{goldspink2010modelling, salganik2006experimental} and the inherent prediction ceilings for stochastic phenomena \cite{conte2012computational, yazdi2024computational}.

\noindent \textbf{\textit{Our Rebuttal:}} The inherent complexity and stochasticity of human behavior do not excuse unreliable simulations; on the contrary, they demand more rigorous methodologies. Our objective is not the perfect replication of an elusive "ground truth," but the creation of simulations demonstrably faithful to observed human behavioral patterns. This very complexity underscores the critical need for the robust validation and reliability-improving frameworks we advocate, ensuring that simulations, even if imperfect, are rendered as trustworthy and insightful as possible.

\section{Conclusion}
We argue that current LLM-based human simulations remain unreliable, while highlighting the key challenges that underlie this limitation. To address these issues, we outline a pathway forward centered on enriched data sources, advances in LLM modeling, and verifiable design principles. These proposed directions aim to substantially enhance the reliability of LLM-based simulations. We also rebut alternative views to highlight the necessity of moving beyond pragmatic compromises toward a higher standard of verifiable reliability. As the volume of simulation studies continues to grow rapidly, our framework serves as a timely call to action for the LLM research community to pursue more rigorous, reliable, and validated simulation methodologies.


\section*{Limitations}

\noindent \textbf{Risks of Misinformation and Economic Implications.}
The potential misuse of highly realistic simulations raises concerns about the spread of misinformation and the erosion of trust in digital interactions. From an economic standpoint, while simulations can reduce costs for behavioral research and training data generation, as shown in Section~\ref{sec:solutions}, there is a risk of creating self-reinforcing feedback loops if synthetic data contaminates future model training.

\noindent \textbf{Costs of Simulation Construction.}
While the proposed framework prioritizes establishing theoretical and methodological foundations for LLM-based human simulation, future empirical work will need to rigorously evaluate the trade-offs between computational costs and performance gains. Initial efforts could prioritize bias mitigation, include multi-modal data sources and implement reflection mechanisms. 
Moreover, the reliance on LLMs for human simulation incurs significant API costs. Although cheaper alternatives are available, they often lack generalizability. A promising approach involves integrating specialized LLMs tailored for specific tasks, utilizing models that excel in particular domains while ensuring efficient communication and data sharing between the models.

\noindent \textbf{Ethical and Social Considerations.} The data collection methods proposed in Section~\ref{sec:solutions}, which involve wearable sensors, present practical challenges related to privacy protection and ethical data usage. Key concerns include obtaining participant consent and ensuring the security of the collected data, both of which pose significant challenges.

\bibliography{neurips_2024}
\appendix

\end{document}